\documentclass[twoside,11pt]{article}

\usepackage{jmlr2e}
\usepackage{amsmath}
\usepackage{subfig}

\newcommand{\vecx}{\mbox{\boldmath$x$}}
\newcommand{\vecy}{\mbox{\boldmath$y$}}

\newcommand{\vece}{\mbox{\boldmath$e$}}

\newcommand{\matW}{\mbox{\boldmath$W$}}

\newcommand{\matO}{\mbox{\boldmath$O$}}
\newcommand{\matX}{\mbox{\boldmath$X$}}
\newcommand{\matU}{\mbox{\boldmath$U$}}
\newcommand{\matI}{\mbox{\boldmath$I$}}

\newcommand{\matK}{\mbox{\boldmath$K$}}

\newcommand{\matA}{\mbox{\boldmath$A$}}

\newcommand{\matB}{\mbox{\boldmath$B$}}
\newcommand{\matS}{\mbox{\boldmath$S$}}
\newcommand{\matY}{\mbox{\boldmath$Y$}}

\newcommand{\pv}{\mbox{\boldmath$\phi$}}
\newcommand{\xv}{\mbox{\boldmath$\xi$}}
\newcommand{\Pv}{\mbox{\boldmath$\Phi$}}
\newcommand{\bv}{\mbox{\boldmath$\beta$}}

\newcommand{\bx}{\vecx}

\jmlrheading{1}{2013}{1-48}{4/00}{10/00}{James Barrett and Anthony C.\ C.\ Coolen}

\ShortHeadings{Dimensionality Detection and Data Integration via the GP-LVM}{James Barrett and Anthony C.\ C.\ Coolen}
\firstpageno{1}

\begin{document}

\title{Dimensionality Detection and Integration of Multiple Data Sources via the GP-LVM}

\author{\name James Barrett \email james.j.barrett@kcl.ac.uk \\
		\name Anthony C.\ C.\ Coolen \email ton.coolen@kcl.ac.uk\\
       \addr Institute for Mathematical and Molecular Biomedicine\\
       King's College London\\
       London, SE1 1UL, U.K.}

\editor{Leslie Pack Kaelbling}

\maketitle

\begin{abstract}
The Gaussian Process Latent Variable Model (GP-LVM) is a non-linear probabilistic method of embedding a high dimensional dataset in terms low dimensional `latent' variables. In this paper we illustrate that maximum a posteriori (MAP) estimation of the latent variables and hyperparameters can be used for model selection and hence we can determine the optimal number or latent variables and the most appropriate model. This is an alternative to the variational approaches developed recently and may be useful when we want to use a non-Gaussian prior or kernel functions that don't have automatic relevance determination (ARD) parameters. Using a second order expansion of the latent variable posterior we can marginalise the latent variables and obtain an estimate for the hyperparameter posterior. Secondly, we use the GP-LVM to integrate multiple data sources by simultaneously embedding them in terms of common latent variables. We present results from synthetic data to illustrate the successful detection and retrieval of low dimensional structure from high dimensional data. We demonstrate that the integration of multiple data sources leads to more robust performance. Finally, we show that when the data are used for binary classification tasks we can attain a significant gain in prediction accuracy when the low dimensional representation is used.
\end{abstract}

\begin{keywords}
  dimensionality reduction, data integration, gaussian process latent variable model, hyperparameter optimisation 
\end{keywords}

\section{Introduction}

Generating a low dimensional representation of a dataset can be an effective way to eliminate redundancy and to better study any structure in the data and extract relevant information. In high dimensional data the number of features greatly outnumbers the samples available. Consequently, when we attempt to use the data for classification or regression tasks we are prone to the problem of overfitting. This occurs when data analysis methods tend to perform well while training on the data but performance on unseen samples can be significantly lower. Our aim in this paper is to reduce the risk of overfitting by using a low dimensional representation of the data.

In order to do this effectively we would like to know what the optimal dimension of the latent variable space is. Too few latent variables may fail to capture all of the relevant structure in a dataset. Using too many may still leave us vulnerable to avoidable overfitting. In the original GP-LVM it is not clear how many latent variables should be used, nor which type of kernel function is most appropriate to best model the relationship between the high and low dimensional data.

Recent advances in sparse Gaussian Process (GP) regression and variational GP methods \citep{TIT09} have been successfully applied to the GP-LVM \citep{LAW10}. A variational lower bound on the posterior over latent variables is maximised with respect to the variational parameters and model hyperparameters. Unnecessary latent dimensions are effectively `turned off' by sending the corresponding ARD hyperparameters to zero. The true dimensionality is estimated as the number of non-zero ARD hyperparameters.

We illustrate in this paper that MAP estimation can be used as an alternative. To do this we construct a Gaussian approximation of the posterior distribution over latent variables which we then integrate over to give an approximation of the hyperparameter posterior. The approximated hyperparameter posterior is used to determine the optimal hyperparameters, intrinsic dimensionality and the overall model likelihood.

There have been several extensions of the GP-LVM that depend upon a non-Gaussian prior over the latent variables. A rank prior is used in \citet{GEI09} to force the GP-LVM to use as few latent dimensions as possible. In the Discriminative GP-LVM \citep{URT07} a prior that encourages samples belonging to the same classes to be close together in the latent variable space is used. The variational approach depends on minimising the Kullback-Leibler divergence between the prior distribution and the variational distribution over latent variables. When the prior is non-Gaussian it may no longer be possible to calculate this analytically and impractical to estimate numerically. In these cases the MAP estimate is more flexible. If we don't want to use kernels with ARD hyperparameters then our method still allows for model selection.

A second problem that is becoming increasingly relevant, particularly in biomedical research, is how to integrate or combine multiple sources of data. If two or more datasets complement each other and share common structure we would expect to extract to extract information more robustly if we could simultaneously use all of the data available. The problem is compounded when different sources have very different dimensions. For example, we may have thousands of variables from genome studies but only ten or twenty from imaging experiments. Any attempt to combine both sources will have to address this dimensionality imbalance since there is a risk the high dimensional data will dominate any analysis due to their greater number.

For this purpose we use the GP-LVM to express each dataset in terms of the same latent variables. We allow each source to have different kernel functions with separate hyperparameters. By seeking a common embedding of multiple datasets in terms of shared low dimensional latent variables we can overcome the issue of dimensionality mismatch while simultaneously extracting structure that is common to both datasets.

This is similar to the Shared GP-LVM \citep{SHO05,EK08} which embed multiple observation spaces in terms of common latent variables. The model is used to predict observations in one space given observations in another. A variational version of this model has also been developed \citep{DAM12}. Our aim differs in that we want to improve regression or classification performance by reducing overfitting. A similar idea was used in the Supervised GP-LVM \citep{GAO11} which models both observed data and output data in terms of common latent variables.

In Section 2 we give an overview of the theoretical background to our model and discuss some of the implementational details. In Section 3 we present results from synthetic data that illustrate the performance of the model under various conditions. We also investigate the integration of multiple sources that have different dimensions.

\section{Gaussian Process Latent Variable Model}

Suppose we observe $S$ datasets $\matY_1\in\mathbb{R}^{N\times d_1},\ldots,\matY_S\in\mathbb{R}^{N\times d_S}$. It is assumed each column of $\matY_s$ is normalised to zero mean and unit variance. We assume that these data can be represented in terms of $q$ latent variables where $q<\min_s(d_s)$. 

\subsection{Model Definition}

We can write each data source in terms of common latent variables
\begin{equation}
y_{i\mu}^s = \sum_{m=1}^Mw^s_{\mu m}\phi^s_m(\vecx_i) + \xi^s_{i\mu}
\label{model}
\end{equation}

Where $\phi^s_m:\mathbb{R}^q\to\mathbb{R}^M$ are non-linear mappings that have yet to be specified but themselves may depend on hyperparameters $\pv_s$, the coefficients $w^s_{\mu m}$ map $\phi^s_m(\vecx_i)$ to the observed data space and $\xi^s_{i\mu}$ are noise variables.

Prior distributions $p(\matW_s)$ and $p(\xv_s|\beta_s)$ are assumed to be zero mean Gaussians with covariances given by $\big<w^s_{\mu m}w^{s'}_{\nu n}\big> = \delta_{ss'}\delta_{\mu\nu}\delta_{mn}$ and $\big<\xi^s_{i\mu}\xi^{s'}_{j\nu}\big> = \beta_s^{-1}\delta_{ss'}\delta_{ij}\delta_{\mu\nu}$ respectively. For notational simplicity we define $\bv = \{\beta_1,\ldots,\beta_S\}$, $\Pv = \{\pv_1,\ldots,\pv_S\}$, $\matW = \{\matW_1,\ldots,\matW_S\}$, $\xv=\{\xv_1,\ldots,\xv_S\}$ and $\matY = \{\matY_1,\ldots,\matY_S\}$. The data likelihood factorises over samples
\[
p(\matY|\matX,\matW,\xv,\bv,\Pv) = \prod_{i=1}^N p(\vecy_i|\vecx_i,\matW,\xv_i,\bv,\Pv)
\]
where $p(\vecy_i|\vecx_i,\matW,\xv_i,\bv,\Pv)$ is also a Gaussian distribution with mean $\matW\vecx_i$ and covariance $\big<y^s_{i\mu}y^{s'}_{j\nu}\big> = \beta^{-1}\delta_{ss'}\delta_{ij}\delta_{\mu\nu}$. As in the original GP-LVM we marginalise $\matW$ and $\xv$ to get the data likelihood. Since $\matW$ and $\xv$ are Gaussian distributed it followed that $\matY$ is also Gaussian with mean $\left<y_{i\mu}\right>=0$ and covariance
\begin{align*}
\big<y^s_{i\mu}y^{s'}_{j\nu}\big>&=\delta_{ss'}\delta_{\mu\nu}\left(\sum_m\phi^s_m(\vecx_i)\phi^s_m(\vecx_j) + \beta^{-1}_s\delta_{ij}\right)\\
&=\delta_{ss'}\delta_{\mu\nu}K_s(\vecx_i,\vecx_j)
\end{align*}
where $K_s(\vecx_i,\vecx_j) = \sum\phi_m(\vecx_i)\phi_m(\vecx_j) +\beta^{-1}_s\delta_{ij}$ is called the kernel matrix. The data likelihood can then be written as
\begin{equation}
p(\matY|\matX,\bv,\Pv) =\prod_{s=1}^S\prod_{\mu=1}^{d_s} \frac{e^{-\frac{1}{2}\small\vecy^s_{:,\mu}\matK^{-1}_s\vecy^s_{:,\mu}}}{(2\pi)^{\frac{N}{2}}|\matK_s|^{\frac{1}{2}}}
\label{datalikelihood}
\end{equation}
where $\vecy_{:,\mu}$ is the $\mu$th column of $\matY$. This can be interpreted as a product of $d_s$ Gaussian Processes \citep{RAS06} per source, each one mapping the shared latent variables onto the $d_s$ covariates for that source. Since each source can be modelled using different non-linear kernels with different hyper-parameters the GP-LVM offers a highly flexible framework for integrating diverse data sources.

\subsection{Hyperparameter Optimisation and Model Selection}
\label{modelselection}
Following the Bayesian formalism we can specify three levels of uncertainty at which we wish to infer certain quantities, having observed the data $\matY$:
\begin{itemize}
\item
Microscopic parameters: $\{\matX\}$
\item
Hyperparameters: $\{\bv,\Pv\}$
\item
Models: $H = \{q,\phi_m\}$
\end{itemize}
A choice of model amounts to selecting the dimension of the latent variable space, $q$, and a choice of which kernel function we want to use. We can then specify posterior distributions over the quantities we wish to infer
\begin{align}
p(\matX|\matY,\bv,\Pv,H) &= \frac{p(\matY|\matX,\bv,\Pv,H)p(\matX|H)}{\int \text{d}\matX^{\prime} p(\matY|\matX^{\prime},\bv,\Pv,H)p(\matX^{\prime}|H)}\label{Xlikelihood}\\
p(\bv,\Pv|\matY,H) &= \frac{p(\matY|\bv,\Pv,H)p(\bv,\Pv|H)}{\int \text{d}\bv^{\prime} \text{d}\Pv^{\prime} \,p(\matY|\bv^{\prime},\Pv^{\prime},H)p(\bv^{\prime},\Pv^{\prime}|H)}\label{hyp_posterior}\\
P(H|\matY) &= \frac{p(\matY|H)p(H)}{\sum_{H^{\prime}}p(\matY|H^{\prime})p(H^{\prime})},\nonumber
\end{align}
where
\begin{align}
p(\matY|\bv,\Pv,H)&= \int \text{d}\matX p(\matY|\matX,\bv,\Pv,H)p(\matX|H)\label{intractableX}\\
p(\matY|H) &= \int \text{d}\bv \text{d}\Pv \,p(\matY|\bv,\Pv,H)p(\bv,\Pv|H).\nonumber
\end{align}
To find the optimal latent variable representation, $\matX^{\star}$ we will numerically minimise the negative log of of (\ref{Xlikelihood}) with respect to $\matX$. We choose a flat, improper prior for $\matX$. Ignoring any constants, the function to minimise is then given by
\begin{equation}
\mathcal{L}_X(\matX;\bv,\Pv) = \sum_s \left[\frac{d_s}{2N}\text{tr}(\matK^{-1}_s\matS_s) + \frac{d_s}{2N}\log |\matK_s| +  \frac{d_s}{2}\log 2\pi\right]
\label{LLx_multiple}
\end{equation}
where $\matS_s = \frac{1}{d_s}\matY_s\matY_s^T$. It has been proposed by \citet{URT06} and \citet{WANG08} that each source should be rescaled by $d_{tot}/d_s$ where $d_{tot} = \sum_s d_s$. This is equivalent to raising the contribution of each source in (\ref{datalikelihood}) to the power of $d_{tot}/d_s$ and effectively regards each source as having the same dimensionality. Experimental evidence indicated that this was unnecessary provided that the hyperparameters for each source are optimised.

\subsection{Optimisation of hyperparameters}
\label{hypopt}
Do determine the optimal hyperparameters we would ideally maximise (\ref{hyp_posterior}). However, the integral (\ref{intractableX}) is both analytically and numerically intractable. One option is to simply optimise $\mathcal{L}_X(\matX;\bv,\Pv)$ with respect to $\bv$ and $\Pv$ as well as $\matX$. Optimisation alternates between $\matX$ and the hyperparameter until a solution if converged on. This is what is done in the original GP-LVM.

Within the Bayesian formalism this is equivalent to assuming $p(\matX|\matY,\bv,\Pv,H)=\delta(\matX - \matX^{\star})$, where $\matX^{\star} = \min_X \mathcal{L}_X$, such that the integral in (\ref{intractableX}) is equal to $p(\matY|\matX^{\star},\bv,\Pv,H)p(\bv,\Pv|H)$. In other words, when optimising the hyperparameters no uncertainty as to the true location of $\matX$ is taken into account.

A slightly better approximation is to expand the likelihood function (\ref{LLx_multiple}) to second order around the minimum $\matX^{\star}$ to get
\begin{equation}
\mathcal{L}_X(\matX;\bv,\Pv) \approx \mathcal{L}_X(\matX^{\star};\bv,\Pv) +\frac{1}{2} \sum_{i,j}^{N}\sum_{\mu,\nu}^{q}(x_{i\mu}^{\star} - x_{i\mu})(x_{j\nu}^{\star} - x_{j\nu})A_{i\mu,j\nu}
\end{equation}
where
\begin{equation}
A_{i\mu,j\nu} =\frac{\partial^2}{\partial x_{i\mu} \partial x_{j\nu}}\mathcal{L}_X(\matX;\bv,\Pv\})\bigg|_{\matX =\matX^{\star}}
\end{equation}
\begin{align*}
p(\matY|\bv,\Pv,H) &= \int \text{d}\matX e^{-N\mathcal{L}_X(\small\matX;\bv,\Pv)}\\
&=p(\matY|\matX^{\star},\bv,\Pv,H)\int \text{d}\matX e^{-\frac{1}{2}\sum_{ij}\sum_{\mu\nu}(x_{i\mu}^{\star} - x_{i\mu})(x_{j\nu}^{\star} - x_{j\nu})A_{i\mu,j\nu}}\\
&=p(\matY|\matX^{\star},\bv,\Pv,H)(2\pi)^{Nq/2}|\matA(\matX^{\star},\bv,\Pv)|^{-1/2}
\end{align*} 
Second order partial derivatives are given in Appendix \ref{partial_derivatives}. The optimal hyperparameters are determined by minimising
\begin{align}
\mathcal{L}_{hyp}(\bv,\Pv) &= \mathcal{L}_X(\matX^{\star};\bv,\Pv) +\frac{1}{2N}\log|\matA(\matX^{\star},\bv,\Pv)|-\frac{q}{2}\log 2\pi
\label{LLhyp_multiple}
\end{align}
The hyperparameter posterior $p(\bv,\Pv|D)$ does not factorise over sources since the determinant of $\matA$ is a product of a sum of partial derivatives for each source.

\subsection{Invariance under Unitary Transformations}
The kernel functions considered in this paper are all invariant under arbitrary unitary transformations. Let $\matU$ be a unitary matrix, such that $\matU^{\text{T}}\matU = \matU\matU^{\text{T}} = \matI$ and let $\tilde{\vecx} = \matU\vecx$. Then  $\tilde{\vecx}_i\cdot\tilde{\vecx}_j = \vecx_i\matU^{\text{T}}\matU\vecx_j = \vecx_i\cdot\vecx_j$ and  $(\tilde{\vecx}_i-\tilde{\vecx}_j)^2 = (\vecx_i - \vecx_j)\matU^{\text{T}}\matU(\vecx_i - \vecx_j) = (\vecx_i-\vecx_j)^2$. This invariance under unitary transformations induces symmetries in the posterior search space of $\matX \in \mathbb{R}^{N\times q}$.

This leads to two practical problems. Firstly, the solution to minimising $\mathcal{L}_X$ is not unique since we can apply any unitary transformation to $\matX$, for example rotation about the origin or reflection through an axis, we will still have an equally valid solution.

The second problem is that there will exist certain directions in the posterior search space that have zero curvature. Consequently $\log|\matA|$ is not defined since several eigenvalues of $\matA$ will always be zero. It is therefore necessary to eliminate these symmetries.

Rotational symmetry can be broken by selecting preferred directions in the latent variable space and applying a sufficient number of unitary transformations to `pin down' $\matX$. Specifically, given an arbitrary basis $\{\vece_1,\ldots,\vece_q\}$ in $\mathbb{R}^{q}$ we can always find a unitary transformation $\matO_1$ such that $\matO_1\vecx_1 = (\tilde{x}_{11},0,\ldots,0)$. That is, we rotate the first sample until it aligns with the $\vece_1$-axis. Similarly, we can find another transformation $\matO_2$ such that $\matO_2\tilde{\vecx_2} = (\tilde{\tilde{x}}_{21},\tilde{\tilde{x}}_{22},0,\ldots,0)$ with $\matO_2\tilde{\vecx}_1=\tilde{\vecx}_1$. We require $q$ such transformations in total, unless $N<q$ in which case we need only $N$.

In practice this is very simple to implement since we can simply populate the upper right hand corner of $\matX$ with zeros. In total $(q^2-q)/2$ elements are set to zero and $\mathcal{L}_X$ can be considered a function of $Nq - (q^2-q)/2$ variables to optimise over.

To eliminate reflectional symmetries we simply require $\tilde{x}_{11} > 0, \tilde{x}_{22} > 0,\ldots,\tilde{x}_{qq}>0$. This can be achieved straight-forwardly with appropriate reflection matrices. Note that there may not be a unique solution to the optimisation of $\mathcal{L}_X$ if either $|\vecx_1| \approx 0$ (rotational symmetry not broken) or $x_{22} \approx 0$ (reflectional symmetry not broken) etc. However, $\log|\matA|$ will always be well defined.
\subsection{Algorithm}
\label{algorithm}
We have implemented the model in Matlab. Minimisation of the likelihood functions is done using Scaled Conjugate Gradients. This requires the first order partial derivatives (see Appendix \ref{partial_derivatives}).
\subsubsection{A Single Data Source}
Given a single dataset $\matY$ we proceed as follows:
\begin{enumerate}
\item Choose a kernel function, set $q$, $\bv$ and $\Pv$ to initial values.
\item Find $\matX^{\star}= \min\mathcal{L}_X$.
\item Find $\{\beta^{\star},\Pv^{\star}\} = \min \mathcal{L}_{hyp}$.
\item Repeat for different values of $q$.
\item Compare the minimum value obtained for $\mathcal{L}_{hyp}$ for each value of $q$ in order to determine which value of $q$ is optimal.
\item Repeat for different kernel functions and compare the minimum values to $\mathcal{L}_{hyp}$ to determine which type is optimal.
\end{enumerate}

Note that the minimisation of $\mathcal{L}_X$ is made difficult by the existence of several local minima whenever a non-linear kernel function is used. To try and locate the global minimum we can make many attempts at minimisation where each attempt starts from a different randomly generated initial point. Computationally, these minimisation attempts can be made in parallel. Alternatively, we can initialise $\matX$ using the first $q$ principal components but this may not lead to the global minimum.

Each evaluation of $\mathcal{L}_{hyp}$ requires $\matX^{\star}$ to be recalculated since it may change for different values of hyperparameters. In general, $\matX^{\star}$ will be shifted by some scale factor and consequently we can use the previous value of $\matX^{\star}$ to initialise the minimisation problem.

\subsubsection{Multiple Data Sources}
Having observed $S$ datasets $\matY_1,\ldots,\matY_S$ we begin by finding the optimal hyperparameters for each source separately. We then minimise $\mathcal{L}_X$ with respect to $\matX$, while holding the hyperparamters for each source fixed. Finally, we evaluate $\mathcal{L}_{hyp}$ at the optimal $\matX$ and optimal hyperparameters. As in the case of a single data source we use the minimum value of $\mathcal{L}_{hyp}$ to determine the optimal value of $q$ and the optimal kernel function.
\section{Application to Synthetic Data}

\begin{figure}[h!]
\centering
\subfloat[`True' latent variables]{\includegraphics[scale =0.8]{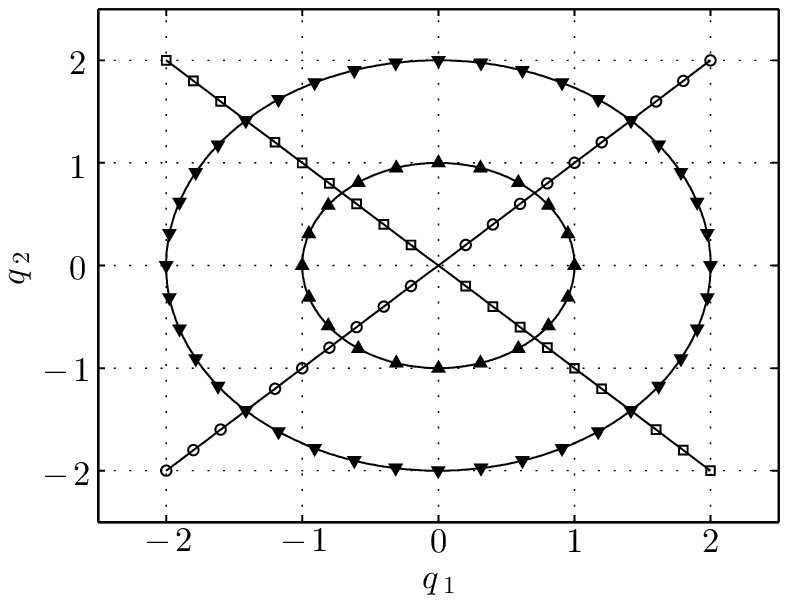}}
\subfloat[Retrieved latent variables]{\includegraphics[scale=0.8]{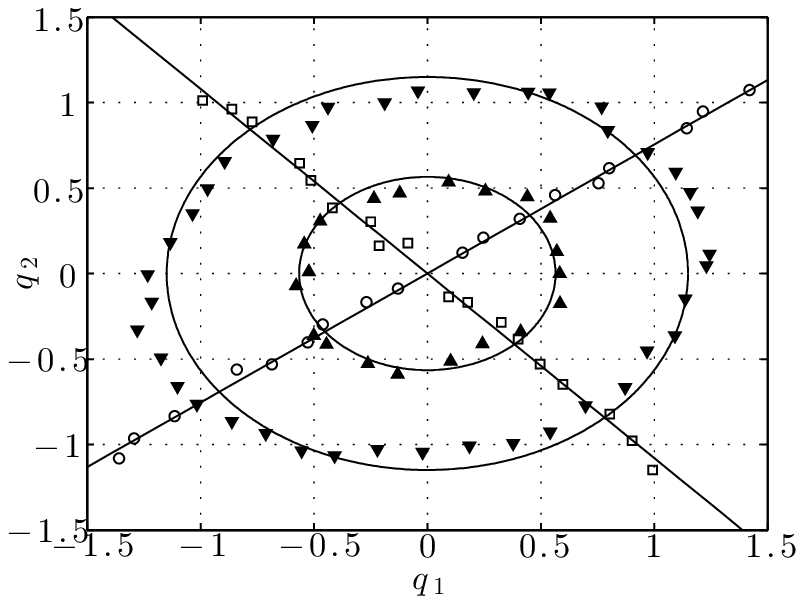}}
\caption{(a) `True' latent variables from which synthetic high dimensional data are generated. The lines serve to guide the eye in qualitatively assessing how well the model extracts the original low dimensional structure. In (b) an example is given of the low dimensional structure retrieved from a synthetic dataset. The underlying structure is clearly visible despite some corruption due to added noise. The errors corresponding to these latent variables can be found in the first row of Table \ref{errors}. Note that the retrieved latent variables are on a different scale compared to that of the originals.}
\label{true_synth}
\end{figure}

In order to generate synthetic data we first construct a matrix of `true' low dimensional data $\matX\in\mathbb{R}^{N\times q}$. We use the specific set of low dimensional data with $q=2$ and $N=96$ shown in Figure \ref{true_synth} (a). These data have a distinctive pattern that allows us to quantitatively and qualitatively assess how well the model performs at retrieving the `true' $\matX$. The matrix $\matX$ is projected into a high dimensional space according to the assumed model (\ref{model}) in order to generate the synthetic high dimensional data $\matY$.

Figure \ref{true_synth} (b) shows the matrix $\matX$ that was retrieved from a synthetic dataset with $d=10$ and $\beta= 0.1$ that was constructed using a linear mapping. It is clear that the original structure has been recovered although it was been slightly corrupted by noise.

We can also define three ad hoc error measures to quantify the quality of the recovered low dimensional data by comparing them to the `true' data. Firstly, samples that belong to either of the two circles should be equidistant from the origin. If $\tilde{r}$ is the mean distance from the origin (of the recovered data) then we can define the mean radial error as
\[
\mathcal{E}_{radial} = \frac{1}{|C|}\sum_{i\in C}\frac{|\vecx_i|-\tilde{r}}{\tilde{r}}
\]
where $C$ is the set of points belonging to the circle and $|C|$ is the number of samples in that set. The errors for both circles are summed.

Similarly, the angle between adjacent samples on either circle should be constant. In the case of the larger circle the angular separation should be $\tilde{\theta} = 2\pi/20$. If we let $\Delta\theta_i$ denote the angle between $\vecx_i$ and the neighbouring point then we can define the mean angular error as
\[
\mathcal{E}_{angular} = \frac{1}{|C|}\sum_{i\in C}\frac{\Delta\theta_i-\tilde{\theta}}{\tilde{\theta}}
\]
For the samples belonging to either line we can try to fit a line by writing $x_2 = \alpha x_1$. The value of $\alpha$ which minimises the sum of squared errors $\sum_i(x_{i2} - \alpha x_{i1})^2$ is given by $\hat{\alpha} = \sum x_{i1}x_{i2}/\sum x_{i1}^2$. We can then define the total sum of squares $SS_{tot} = \sum (x_{i2} - \bar{x}_{i2})^2$ and the sum of squared residuals $SS_{err} = \sum (x_{i2} - \alpha x_{i1})^2$ and finally define
\[
\mathcal{E}_{linear} = \frac{SS_{err}}{SS_{tot}}
\]

These measure have two desirable properties. All three error measures are zero for the `true' low dimensional data. Secondly, they are invariant under rescaling of $\matX$. This is important because the data $\matY$ are normalised to zero mean and unit variance before the algorithm attempts to find $\matX$ so there is no reason to expect the retrieved and the `true' latent variables to have the save overall scale.

\subsubsection*{Dependence on $\beta$ and $d$}

We can test how the error depends on the noise level $\beta$ and dimension $d$. Synthetic data were generated using a linear mapping. As shown in Table \ref{errors} the errors increase as more noise is added to the data whereas the errors decrease when the dimension of the observed dataset is higher.

\begin{table}[h]
  \begin{center}
\subfloat[Dependence on $\beta$]{
    \begin{tabular}{|c|c|c|c|}
    \hline
    $\beta$ & $\mathcal{E}_{radial}$ & $\mathcal{E}_{angular}$ & $\mathcal{E}_{linear}$\\
	\hline
	0.1 &     0.0060  &  0.0046  &  0.0079 \\
	0.5 &     0.0766  &  0.0813  &  0.2577 \\
	1.0 &    0.0998   & 0.1701   & 0.3263 \\
    \hline
    \end{tabular}
    }
    \subfloat[Dependence on $d$]{
        \begin{tabular}{|c|c|c|c|}
    \hline
    $d$ & $\mathcal{E}_{radial}$ & $\mathcal{E}_{angular}$ & $\mathcal{E}_{linear}$\\
	\hline
	10 &         0.0944   & 0.0454  &  0.5491\\
	100 &         0.0061   & 0.0051  &  0.0108 \\
	1000 &       0.0004   & 0.0008  &  0.0016 \\
    \hline
    \end{tabular}
    }
  \end{center}
\caption{(a) The magnitude of the errors increases as more noise is added (for fixed $d$) to the synthetic data. (b) For fixed noise levels the greater $d$ is the better the extraction of the `true' low dimensional structure from a dataset.}
\label{errors}
\end{table}

\subsection{Dimensionality detection and model selection}

We can also compare the likelihood of different models $H_{q,\varphi} = \{q,\varphi_m\}$. This allows us to detect any intrinsic low dimensional structure to a dataset in a probabilistic manner. We can also determine which type of kernel function is most appropriate.

\begin{figure}
\centering
\includegraphics[scale = 0.8]{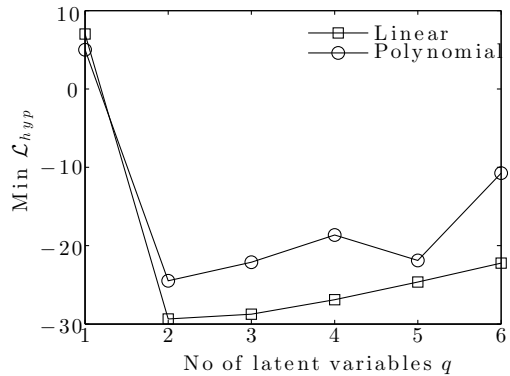}
\caption{Plot of the minimal values of $\mathcal{L}_{hyp}$ obtained for different values of $q$ and two different kernels, the linear kernel and the polynomial kernel. Both the kernel types detect that $q=2$ is the optimal dimension. Furthermore, the model can distinguish that the linear kernel offers the best explanation of the data in this case.}
\label{dimension_detection}
\end{figure}

Figure \ref{dimension_detection} shows the minimum of $\mathcal{L}_{hyp}$ as a function of $q$ using two different kernels. The data was generated with a linear kernel, $d=10$ and $\beta = 0.01$. The model has detected that a linear kernel with $q=2$ is indeed the best explanation of these data.

Note that multiple attempts need to be made to locate the global minimum of $\mathcal{L}_{hyp}$ when using the polynomial kernel. In this case we can check that we have reached the global minimum by visually checking the solution (for $q=2$) and comparing it to the `true' low dimensional data. When $q>2$ it is observed that the first two latent variables are very similar to the `true' solution. The additional latent variables tend to be relatively small in magnitude and presumably are being used by the model to explain some of the noise in the observed data.

We can compute the likelihood ratio between the linear and polynomial models. The ratio is given by $e^{\min\mathcal{L}_{hyp}^{poly}-\min\mathcal{L}_{hyp}^{lin}}$. In this case, when $q=2$, the linear model is approximately 130 times more likely than the polynomial model.

\subsection{Integration of multiple sources}

\begin{figure}[h!]
\centering
\begin{tabular}{c c}
\subfloat[Linear]{\includegraphics[scale = 0.8]{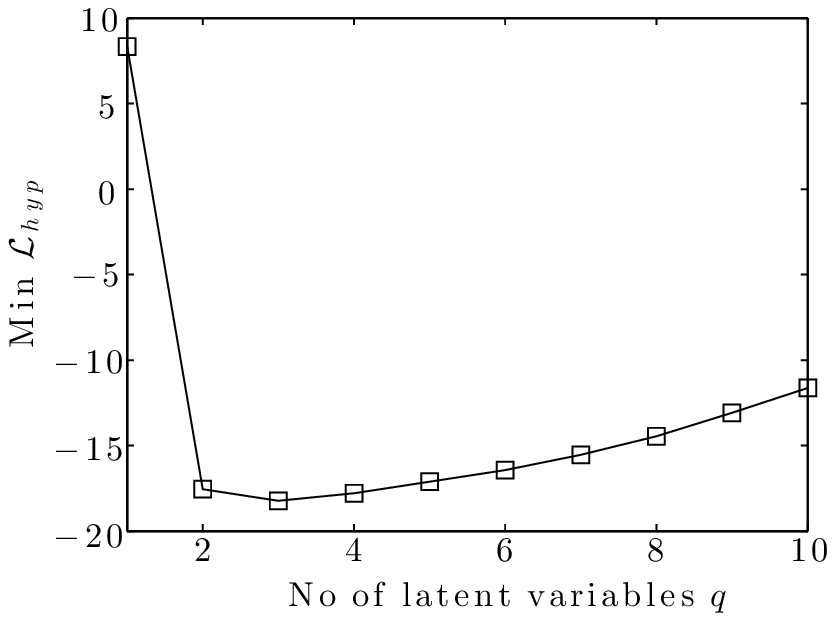}} & \subfloat[Polynomial]{\includegraphics[scale = 0.8]{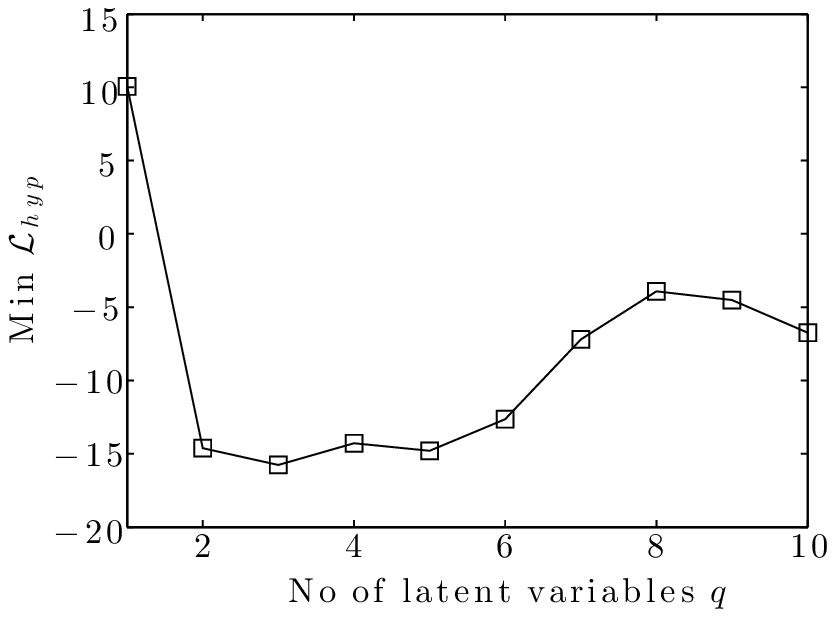}}\\
\multicolumn{2}{c}{\subfloat[Combined]{\includegraphics[scale = 0.8]{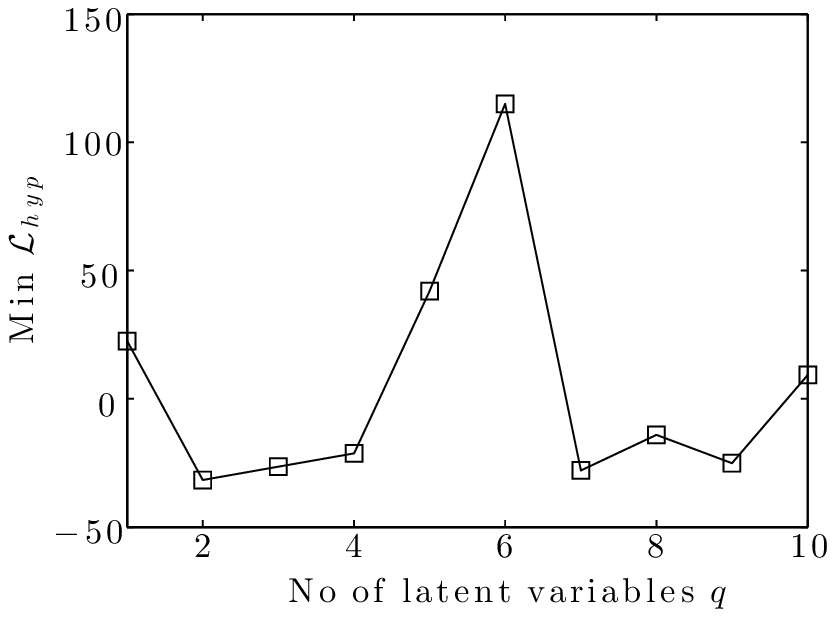}}}\\
\end{tabular}
\caption{Plots of the minimum value of $\mathcal{L}_{hyp}$ obtained for different values of $q$. (a) Results for the `Linear' dataset, with three being identified as the optimal dimension. (b) The `Polynomial' dataset, again with $q=3$ being the optimal number of latent variables. (c) When both sources are combined the model finds $q=2$ is the optimal solution. This illustrates the fact that combined multiple sources of information can lead to a more accurate detection of low dimensional structure.}
\label{integration_plot}
\end{figure}

We can also show that a combination of multiple data sources simultaneously leads to a more robust recovery of the true solution. Two datasets are generated with a linear and a polynomial kernel. The dimensions are $d_{lin}=100$ and $d_{poly} = 10$ and noise levels of $\beta_{lin} = 0.03$ and $\beta_{poly}=0.02$. Taken alone the model detects $q=3$ as the optimal solution for each dataset. When combined however the model detects the correct solution of $q=2$ as shown in Figure \ref{integration_plot}.

Furthermore, the overall model likelihood is higher for the model that combines both sources. We can also see that as well as a minimum at $q=2$ the negative log likelihood has a second pronounced minimum at $q=7$. One interpretation of this is that the model can explain both sources separately when given a sufficient number of latent variables. However the overall minimum at $q=2$ indicates the fact that these sources have a common structure.

Table \ref{integration_error} shows the errors associated with the recovered low dimensional data. In the case of $\mathcal{E}_{linear}$ there is an improvement when both sources are combined. In $\mathcal{E}_{radial}$ and $\mathcal{E}_{angular}$ the error made when both sources are combined is somewhere between the errors made using each source individually.

\begin{table}[h]
  \begin{center}
    \begin{tabular}{|c|c|c|c|}
    \hline
    Kernel & $\mathcal{E}_{radial}$ & $\mathcal{E}_{angular}$ & $\mathcal{E}_{linear}$\\
	\hline
	Combined &         0.00111   &0.00037  & 0.00053\\
	Linear &            0.00416 &  0.00100 &  0.00091 \\
	Poly &           0.00021  &  0.00035  &  0.00086\\
    \hline
    \end{tabular}
  \end{center}
\caption{The errors corresponding to the optimal $q=2$ representation of both datasets and their combination. }
\label{integration_error}
\end{table}

\subsection{Reduction in overfitting}

To illustrate the practical benefit of a latent variable representation in reducing overfitting we now generate low dimensional data, $\matX$, with $q=2$ where each sample belongs to a binary class $\{-1,+1\}$. We choose $N=100$ and $d=100$. We generate 50 samples with a class label of $+1$ from a Gaussian distribution with unit variance and mean $(1,1)$. We then generate 50 samples from the $-1$ class from two unit variance Gaussians centred on $(-\tfrac{1}{2},-\tfrac{1}{2})$ and $(\tfrac{1}{2},-\tfrac{1}{2})$. We then project these data into a high dimensional space according to (\ref{model}) with a linear mapping.

In general these data will not be linearly separable so to perform binary classification we use a Support Vector Machine (SVM) with a Radial Basis Function (RBF) kernel. We use Leave One Out Cross Validation (LOOCV) to determine the optimal SVM parameters, which in this case are the so-called `box constraint' parameter (which controls the degree to which the SVM will try and avoid misclassifications during training) and the length scale in the RBF kernel. We also use LOOCV to compute a training success rate (the average percentage of samples correctly classified during training) and a validation success rate (percentage of validation samples correctly classified after the training phase has been completed).

\begin{table}[h]
  \begin{center}
    \begin{tabular}{|c|c|c|}
    \hline
     & $\matY$ ($d=100$)& $\matX^{\star}$ ($q=2$)\\
	\hline
	Training Success &  86.3\%  & 86.6\%\\
	Validation Success & 74.0\% &  83.0\% \\
    \hline
    \end{tabular}
  \end{center}
\caption{Results from running an SVM binary classifier with LOOCV on the original high dimensional data, $\matY$, and the low dimensional representation, $\matX^{\star}$, obtained from the latent variable model. There is a significant reduction in overfitting in the latent variable space.}
\label{reduction_in_overfitting}
\end{table}

In general, the greater the ratio of samples to features is, the more successfully we will detect genuine patterns in the data. To quantify this we train the classifier on all but one of the samples to compute the training success. After training we present the omitted sample and use the trained SVM to classify it. If the SVM has extracted meaningful patterns from the data during the training then we expect to see a validation success that is greater than chance (50\%).

Table \ref{reduction_in_overfitting} shows an increase in the validation success when classification is performed in the latent variable space. The ratio of samples to features is greatly boosted in the latent variable representation and consequently the SVM is less prone to overfitting.

In order to verify that we are observing a systematic reduction in overfitting we repeat this experiment 300 times. Each time we generate low dimensional data at random, project these into a high dimensional space, and add some randomly generated noise. The mean improvment between the validation success on $\matY$ and the success on $\matX^{\star}$ was found to be 8.7\% with a standard deviation of 4.8\%.

\section{Conclusion}

Our aim is to determine an optimal low dimensional representation of high dimensional data in order to reduce the risk of overfitting by boosting the ratio of samples to features. Our results indicate that the performance of classification algorithms, which attempt to predict an outcome from the observed data, is improved when the latent variable representation is used. Furthermore, our model offers an effective method to combine several data sources simultaneously. Our model can be used as a flexible preprocessing method that diminishes the risk of overfitting and detects any intrinsic low dimensional structure.

In order to capture all relevant information we want to know how many latent variables and which kernels are most appropriate to model given datasets. We illustrate that MAP estimation is an effective and flexible method for determining optimal hyperparameters and performing model comparison.

Using a second order expansion of the latent variable posterior we can marginalise the latent variables and obtain an estimate for the hyperparameter posterior. This expansion requires us to eliminate symmetries in the posterior space that exist due to symmetries in the kernel functions, which we do in a computationally simple manner. This has the added benefit of ensuring a unique latent variable representation.

MAP estimation is a flexible alternative to the variational approaches recently developed. Future work could investigate the optimal dimensionality of datasets with models that incorporate outcome information.


\acks{This work was funded under the European Commission FP7 Imagint Project, EC Grant Agreement no. 259881.}

\appendix

\section{Calculation of partial derivatives}
\label{partial_derivatives}
As discussed in Section \ref{algorithm} the first order partial derivates are required for gradient based minimisation of the likelihood function. In Section \ref{hypopt} we required the second order partial derivatives in order to construct a Gaussian approximation of $\mathcal{L}_X$. The two kernel functions considered here are the linear and polynomial kernels. The derivatives of $\mathcal{L}_X$ are given by 
\begin{equation}
\frac{\partial }{\partial \matX}\mathcal{L}_X=\sum_{s=1}^S\sum_{i,j=1}^N\frac{	\partial \mathcal{L}_X}{\partial \matK_{ij}^s}\frac{\partial \matK_{ij}^s}{\partial \matX}
\label{chainrule}
\end{equation}
where
\begin{equation}
\frac{\partial \mathcal{L}_X}{\partial \matK^s}=-\frac{d_s}{2N}\matK_s^{-1}\matS_s\matK_s^{-1}+\frac{d_s}{2N}\matK_s^{-1}
\label{GLL}
\end{equation}
In what follows we drop the index $s$ for clarity. The following identities are used
\begin{align*}
\frac{\partial |\matK|}{\partial \matK}&=|\matK|\matK^{-1}\\
\frac{\partial \text{tr}(\matA\matK^{-1}\matB)}{\partial \matK}&=-(\matK^{-1}\matB\matA\matK^{-1})^{\text{T}}
\end{align*}

\subsection{Linear Kernel}

The kernel is defined by $K(\vecx_i,\vecx_j) = \vecx_i\cdot\vecx_j + \beta^{-1}\delta_{ij}$. First order partial derivatives are
\begin{equation*}
\frac{\partial}{\partial \matX}\mathcal{L}_X=-\frac{d}{N}\matK^{-1}\matS\matK^{-1}\matX+\frac{d}{N}\matK^{-1}\matX 
\end{equation*}

Second order partial derivatives are
\begin{align*}
\frac{\partial}{\partial x_{p\nu}} (-\matK^{-1}\matS\matK^{-1}\matX)_{r\mu}=&-(\matK^{-1}\matS\matK^{-1})_{rp}\delta_{\mu\nu}+(\matK^{-1}\matS\matK^{-1}\matX)_{p\mu}(\matK^{-1}\matX)_{r\nu} +\nonumber\\
&+(\matK^{-1}\matS\matK^{-1}\matX)_{r\nu}(\matK^{-1}\matX)_{p\mu}+(\matK^{-1}\matS\matK^{-1})_{rp}(\matX^T\matK^{-1}\matX)_{\nu\mu} +\nonumber\\
&+(\matX^T\matK^{-1}\matS\matK^{-1}\matX)_{\nu\mu}(\matK^{-1})_{rp}
\end{align*}
\begin{align*}
\frac{\partial}{\partial x_{p\nu}} (\matK^{-1})_{r\mu}=& (\matK^{-1})_{rp}\delta_{\mu\nu}-(\matK^{-1}\matX)_{r\nu}(\matK^{-1}\matX)_{p\mu}-(\matX^T\matK^{-1}\matX)_{\nu\mu}(\matK^{-1})_{rp}
\end{align*}

\subsection{Polynomial Kernel}

Using a binomial expansion the kernel function can be written as
\[
K(\vecx_i,\vecx_j) = \sum_{n=0}^{\alpha} \binom{\alpha}{n}(\bx_i\cdot\bx_j)^n +\beta^{-1} \delta_{ij}
\]
We assume $\alpha=2$. The partial derivatives of $\matK$ with respect to $\matX$ are given by
\begin{align*}
\frac{\partial K_{ij}}{\partial x_{r\mu}} &= 0\\
\frac{\partial K_{ir}}{\partial x_{r\mu}}&= \frac{\partial K_{ri}}{\partial x_{r\mu}} = 2x_{i\mu}(1+\bx_i\cdot\bx_r)\\
\frac{\partial K_{rr}}{\partial x_{r\mu}}&=4x_{r\mu}(1+\bx_r\cdot\bx_r)
\end{align*}
Inserting into (\ref{chainrule}) gives
\begin{equation}
\frac{\partial L}{\partial x_{r\mu}} = 2\sum_{i=1}^N\frac{\partial L}{\partial K_{ir}}2x_{i\mu}(1+\bx_i\cdot\bx_r)
\label{firstorderpolypartial}
\end{equation}
Differentiating (\ref{firstorderpolypartial}) a second time, and using (\ref{GLL}), we get
\begin{align}
&2\sum_i^N\left\{\frac{\partial}{\partial x_{p\nu}}\left[-\frac{d}{2N}\matK^{-1}\matS\matK^{-1} + \frac{d}{2N}\matK^{-1}\right]_{ir}2x_{i\mu}(1+\bx_i\cdot\bx_r)\right.\nonumber\\
&\qquad\qquad+\left.\left[-\frac{d}{2N}\matK^{-1}\matS\matK^{-1} + \frac{d}{2N}\matK^{-1}\right]_{ir}\frac{\partial}{ \partial x_{p\nu}}\big[2x_{i\mu}(1+\bx_i\cdot\bx_r)\big]\right\}
\label{secondorderpolypartial}
\end{align}
The first term inside the fist square brackets is
\begin{align*}
\frac{\partial}{\partial x_{p\nu}}(\matK^{-1}\matS\matK^{-1})_{ir} &= \frac{\partial}{\partial x_{r\mu}}\sum_{tl}\matK^{-1}_{it}\matS_{tl}\matK^{-1}_{lr}\nonumber\\
& = \sum_{tl}\matK^{-1}_{it}\matS_{tl}\left[\frac{\partial}{\partial x_{p\nu}}\matK^{-1}_{lr}\right]+\sum_{tl}\left[\frac{\partial}{\partial x_{p\nu}}\matK^{-1}_{it}\right]\matS_{tl}\matK^{-1}_{lr}
\end{align*}
This can be simplified to
\begin{align*}
\frac{\partial}{\partial x_{r\mu}}(\matK^{-1}\matS\matK^{-1})_{ir} = &\sum_{k=1}^N2\sigma x_{k\nu}(1+\vecx_k\vecx_p))\Big( -[\matK^{-1}\matS\matK^{-1}]_{ik}\matK^{-1}_{pr} -[\matK^{-1}\matS\matK^{-1}]_{ip}\matK^{-1}_{kr} \Big.\nonumber\\
&\qquad\qquad\qquad\Big.- [\matK^{-1}\matS\matK^{-1}]_{pr}\matK^{-1}_{ik}-[\matK^{-1}\matS\matK^{-1}]_{kr}\matK^{-1}_{ip}\Big)
\end{align*}
The second term inside the first square brackets of (\ref{secondorderpolypartial}) is given by
\begin{equation*}
\frac{\partial}{\partial x_{p\nu}}(\matK^{-1})_{ir} = \sum_{k=1}^N(-\matK^{-1}_{ik}\matK^{-1}_{pr} -\matK^{-1}_{ip}\matK^{-1}_{kr})(2\sigma x_{k\nu}(1+\vecx_k\vecx_p))\label{polypartial2}
\end{equation*}
Finally, the second term in (\ref{secondorderpolypartial}) is given by
\begin{align*}
\left[-\frac{d}{2N}\matK^{-1}\matS\matK^{-1} + \frac{d}{2N}\matK^{-1}\right]_{ip}2\sigma x_{i\mu}x_{i\nu}&\qquad\text{when $i\neq p$ and $r = p$}\\
\left[-\frac{d}{2N}\matK^{-1}\matS\matK^{-1} + \frac{d}{2N}\matK^{-1}\right]_{pp}2\sigma\left(\delta_{\mu\nu}(1+\vecx_p^2) + 2x_{p\mu}x_{p\nu}\right)&\qquad\text{when $i=p$ and $r = p$}\\
\left[-\frac{d}{2N}\matK^{-1}\matS\matK^{-1} + \frac{d}{2N}\matK^{-1}\right]_{pr}2\sigma\left(\delta_{\mu\nu}(1+\vecx_p\vecx_r) + x_{p\mu}x_{r\nu}\right)&\qquad\text{when $i=p$ and $r\neq p$}
\end{align*}
and zero when $i\neq p$ and $r \neq p$.

\vskip 0.2in

\bibliography{final_references}

\end{document}